\pdfoutput=1

\documentclass[11pt]{article}

\usepackage[final]{acl}

\usepackage{times}

\usepackage{latexsym}
\usepackage{tabularx}
\usepackage{makecell}

\usepackage{graphicx}     
\usepackage{colortbl}     
\usepackage{makecell}     
\usepackage{booktabs}     
\usepackage{placeins}     

\usepackage[most]{tcolorbox}

\usepackage[T1]{fontenc}

\usepackage[utf8]{inputenc}

\usepackage{microtype}

\usepackage{inconsolata}

\usepackage{graphicx}

\usepackage{amsmath}
\usepackage{amsfonts}
\usepackage{booktabs}
\usepackage{multirow}
\usepackage{colortbl}

\title{CDF-RAG: Causal Dynamic Feedback for Adaptive Retrieval-Augmented Generation}

\author{
  Elahe Khatibi\textsuperscript{*} \quad Ziyu Wang\textsuperscript{*} \quad Amir M. Rahmani \\
  University of California, Irvine, USA \\
  {\tt \{ekhatibi, ziyuw31, a.rahmani\}@uci.edu}
}

\begin{document}
\maketitle
\def\thefootnote{*}\footnotetext{Ziyu Wang and Elahe Khatibi contributed equally to this work.}

\pagestyle{empty}

\begin{abstract}
Retrieval-Augmented Generation (RAG) has significantly enhanced large language models (LLMs) in knowledge-intensive tasks by incorporating external knowledge retrieval. However, existing RAG frameworks primarily rely on semantic similarity and correlation-driven retrieval, limiting their ability to distinguish true causal relationships from spurious associations. This results in responses that may be factually grounded but fail to establish cause-and-effect mechanisms, leading to incomplete or misleading insights. To address this issue, we introduce \textbf{C}ausal \textbf{D}ynamic \textbf{F}eedback for Adaptive \textbf{R}etrieval-\textbf{A}ugmented \textbf{G}eneration (CDF-RAG), a framework designed to improve causal consistency, factual accuracy, and explainability in generative reasoning. CDF-RAG iteratively refines queries, retrieves structured causal graphs, and enables multi-hop causal reasoning across interconnected knowledge sources. Additionally, it validates responses against causal pathways, ensuring logically coherent and factually grounded outputs. We evaluate CDF-RAG on four diverse datasets, demonstrating its ability to improve response accuracy and causal correctness over existing RAG-based methods. Our code is publicly available at \url{https://github.com/elakhatibi/CDF-RAG}.
 
\end{abstract}

\section{Introduction}

\begin{figure*}[ht]
    \centering
    \includegraphics[width=0.7\textwidth]{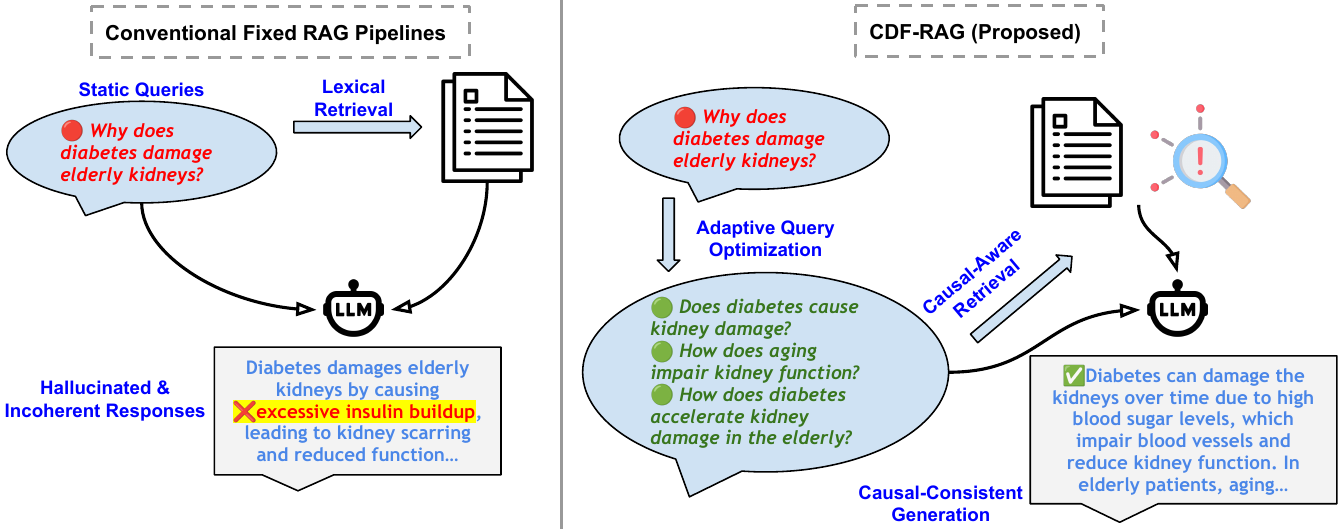}
    \caption{\textbf{Rethinking Retrieval-Augmented Generation (RAG).} (a) Traditional RAG pipelines rely on static queries and keyword- or similarity-based retrieval, often retrieving topically related but causally irrelevant content, which can result in hallucinated or incoherent outputs. (b) \textbf{CDF-RAG} addresses these limitations through reinforcement learning-based query refinement, dual-path retrieval combining semantic vector search with causal graph traversal, and causal-consistent generation, leading to improved factuality and reasoning.}
    \label{fig:cdf-rag-comparison}
\end{figure*}

Large language models (LLMs) such as GPT-4~\cite{achiam2023gpt}, DeepSeek~\cite{liu2024deepseek}, and LLaMA~\cite{touvron2023llama} have demonstrated strong performance across a range of reasoning tasks, including fact-based question answering~\cite{liang2022holistic}, commonsense inference~\cite{huang2019cosmos}, and multi-hop retrieval~\cite{yang2018hotpotqa, zhuang2024efficientrag}. Retrieval-Augmented Generation (RAG)~\cite{lewis2020retrieval} has been introduced to enhance LLMs by retrieving external documents, thereby improving response reliability in knowledge-intensive tasks~\cite{wei2024instructrag, li2024rag}. However, conventional RAG pipelines typically rely on static queries and semantic similarity-based retrieval, which prioritize topically relevant documents rather than those that provide explanatory or causal insights~\cite{jiang2024reasoning, chi2024unveiling}. While effective for shallow fact recall, these strategies often fall short in tasks requiring multi-step causal reasoning~\cite{vashishtha2310causal, jin2023can}.

This reliance on correlation-driven retrieval introduces key challenges for causality-aware reasoning. Traditional RAG systems (as shown in Figure~\ref{fig:cdf-rag-comparison}) struggle to distinguish between statistical associations and true causal relationships~\cite{chi2024unveiling}, leading to retrieved evidence that may appear relevant but lacks directional or explanatory depth. Furthermore, LLMs trained on large-scale observational corpora tend to model co-occurrence patterns rather than causal dependencies, making them prone to conflating correlation with causation—especially in the presence of incomplete or ambiguous evidence. These limitations become more pronounced in multi-hop retrieval, where linking causally related pieces of information is essential for producing coherent reasoning chains~\cite{zhuang2024efficientrag}. However, conventional retrieval strategies typically employ flat or lexical matching techniques, which fail to incorporate causal structure, leading to responses that are locally plausible yet globally inconsistent.

Such shortcomings have direct consequences in real-world applications where causal understanding is critical. In medical decision-making, for instance, associating “high BMI” with “heart disease” may be factually accurate but incomplete without identifying mediating factors such as “hypertension” or “insulin resistance.” When causal evidence is sparse or incorrectly retrieved, LLMs often compensate by hallucinating plausible-sounding but unsupported explanations~\cite{sun2024redeep, yu2024auto}, reducing trustworthiness. Additionally, static query formulation prevents models from adapting retrieval based on reasoning gaps, further exacerbating these issues. While recent work has explored structured retrieval~\cite{jin2024long}, multi-hop planning~\cite{ferrando2024know}, and causal graph construction~\cite{samarajeewa2024causalgraphrag}, these approaches address isolated components rather than providing an end-to-end framework for causal reasoning.

Another key challenge in causal question answering is that many user queries in QA are also vague or underspecified, making effective retrieval even more challenging. While methods like RQ-RAG~\cite{chan2024rqrags}, RAG-Gym~\cite{xiong2025raggym}, and SmartRAG~\cite{gao2024smartrag} introduce query refinement or agentic retrieval mechanisms, they lack dynamic adaptation and causal alignment—often retrieving shallow or loosely connected content. This highlights the need for refinement strategies that are explicitly optimized for causal reasoning.

To address these challenges, we propose \textbf{C}ausal \textbf{D}ynamic \textbf{F}eedback for \textbf{R}etrieval-\textbf{A}ugmented \textbf{G}eneration \textbf{(CDF-RAG)}, a novel framework that integrates reinforcement-learned query refinement, multi-hop causal graph retrieval, and alignment-based hallucination detection into a dynamic reasoning loop. These components enable CDF-RAG to retrieve causally relevant evidence and generate logically coherent responses grounded in causal structures. Our experiments on CosmosQA~\cite{huang2019cosmos}, MedQA~\cite{jin2020disease}, MedMCQA~\cite{pmlr-v174-pal22a}, and AdversarialQA~\cite{bartolo2020beat} show that CDF-RAG consistently outperforms standard and refined RAG models~\cite{chan2024rqrags, xiong2025raggym} across key metrics—demonstrating its effectiveness for generating factually consistent and causally coherent responses in complex QA settings-- providing a robust foundation for trustworthy reasoning in real-world applications.

\textbf{Contributions.} Our paper makes the following contributions:
\begin{itemize}
    \item We introduce \textbf{CDF-RAG}, a unified framework that integrates causal query refinement, multi-hop causal graph retrieval, and hallucination detection into a dynamic feedback loop for causality-aware generation.
    \item We demonstrate that our reinforcement learning (RL)-based query rewriting significantly enhances multi-hop causal reasoning and retrieval quality, outperforming prior refinement approaches.
    \item We show that CDF-RAG achieves state-of-the-art performance on four QA benchmarks, with consistent improvements in causal correctness, consistency, and interpretability over existing RAG-based models.
\end{itemize}

\section{CDF-RAG: Causal Dynamic Feedback for RAG}

\begin{figure*}[ht]
    \centering
    \includegraphics[width=0.9\textwidth]{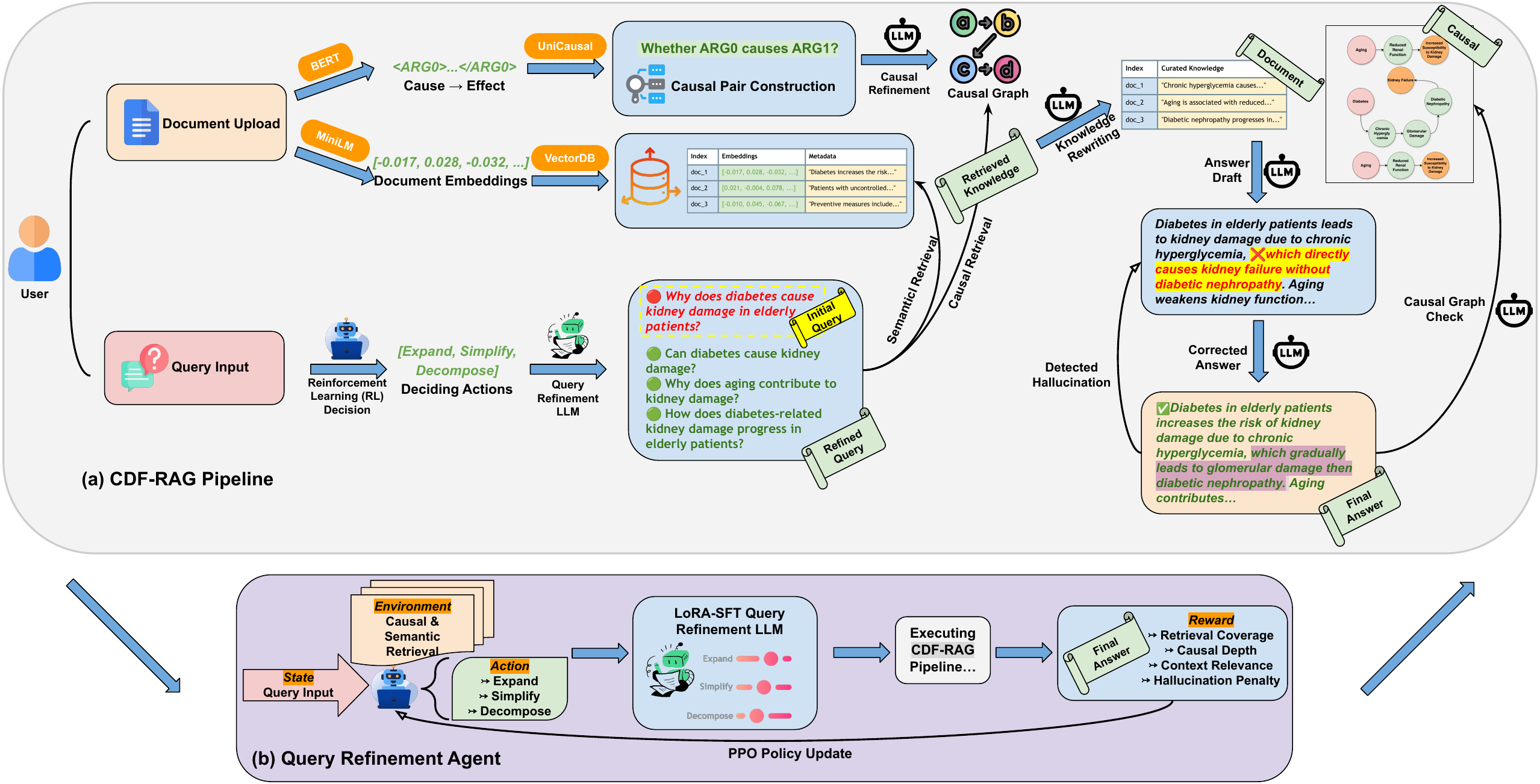}
    \caption{
    \textbf{Overview of CDF-RAG Framework.} (a) The CDF-RAG pipeline refines user queries (LLM + RL), retrieves structured causal and unstructured textual knowledge, applies knowledge rewriting, and ensures factual consistency through causal verification. (b) The PPO-trained query refinement agent optimizes retrieval coverage and causal consistency.
    }
    \label{fig:cdf-rag}
\end{figure*}

We introduce \textbf{CDF-RAG}, a causality-aware extension of RAG. As illustrated in Figure~\ref{fig:cdf-rag}, the system refines user queries via a query refinement LLM trained with RL, retrieves knowledge using a dual-path retrieval mechanism, rewrites knowledge, and applies a causal graph check to ensure factual consistency. By integrating structured causal reasoning, CDF-RAG mitigates hallucinations and enhances interpretability. This approach enables dynamic query adaptation and precise retrieval for causal reasoning tasks. Implementation details can be found in Appendix~\ref{sec:appendix}

\subsection{Causal Knowledge Graph Construction}

CDF-RAG constructs a directed causal knowledge graph $\mathcal{G} = (V, E)$ from textual data to capture causal dependencies beyond correlation. Using UniCausal~\cite{tan2023unicausal}, a BERT-based classifier extracts cause-effect pairs formatted as $C \rightarrow E$, processing annotated inputs \texttt{<ARG0>} and \texttt{<ARG1>} to predict $\hat{y} = g(r_{\text{[CLS]}})$. 

To ensure logical validity, extracted causal pairs are verified by GPT-4 before being encoded into $\mathcal{G}$ as directed triples $(C, E, \text{relation})$. The graph structure enables multi-hop reasoning over causal mechanisms, ensuring retrieved knowledge supports causal inference tasks.

\subsection{Causal Query Refinement via Reinforcement Learning}

Given an initial user query $q$, CDF-RAG applies RL to generate a refined query $\hat{q}$ optimized for causal retrieval. The RL-based query refinement agent models this as a Markov Decision Process (MDP), where the state $s$ represents the query embedding, and the agent selects an action $a \in \{\texttt{expand}, \texttt{simplify}, \texttt{decompose}\}$. Expansion enhances specificity by adding relevant causal factors, simplification removes extraneous details, and decomposition restructures complex queries into atomic subqueries.

The policy $\pi_\theta(a \mid s)$ is initialized via supervised fine-tuning (SFT) on labeled refinement examples:
\[
\mathcal{L}_{\text{SFT}} = -\sum_{t=1}^T \log P_{\phi}(y_t \mid y_{<t}, x)
\]
and further optimized using Proximal Policy Optimization (PPO)~\cite{schulman2017proximal}:


\begin{align*}
\mathcal{L}_{\text{PPO}}(\theta) = \mathbb{E}_t \big[ 
& \min \big( r_t(\theta) \hat{A}_t, \\
& \quad \text{clip}(r_t(\theta), 1 - \epsilon, 1 + \epsilon) \hat{A}_t \big) \big]
\end{align*}
\noindent where $r_t(\theta) = \frac{\pi_\theta(a_t \mid s_t)}{\pi_{\theta_{\text{old}}}(a_t \mid s_t)}$.

The reward function optimizes retrieval effectiveness and causal consistency:


\begin{align*}
R =\; & \lambda_1 \cdot \text{RetrievalCoverage} \\
& + \lambda_2 \cdot \text{CausalDepth} \\
& + \lambda_3 \cdot \text{ContextRelevance} \\
& - \lambda_4 \cdot \text{HallucinationPenalty}
\end{align*}

By refining queries with these criteria, CDF-RAG dynamically adapts retrieval strategies to enhance causal reasoning.

\subsection{Dual-Path Retrieval: Semantic and Causal Reasoning}

To ensure comprehensive and aligned knowledge access, CDF-RAG adopts a dual-path retrieval strategy, integrating semantic vector search with causal graph traversal.

\textbf{Semantic Vector Retrieval.} 
Inspired by dense retrieval methods~\cite{karpukhin2020dense}, we encode the refined query $\hat{q}$ using MiniLM~\cite{wang2020minilm} and perform similarity search in a vector database. This semantic retrieval pathway returns top-$k$ passages $\mathcal{T}_{\text{sem}}$ that offer contextual evidence supporting the query. Unlike sparse retrieval methods such as BM25~\cite{robertson2009probabilistic}, which rely on term frequency heuristics, our approach captures contextual relevance through transformer-based embeddings. This enables richer matching, particularly for lexically divergent yet semantically similar phrases—a common challenge in biomedical and causal reasoning tasks.

\textbf{Causal Graph Traversal.} 
To complement semantic retrieval with structural reasoning, we traverse a domain-specific causal graph $\mathcal{G}$. Given $\hat{q}$, we identify aligned nodes and expand along directed edges to surface causally linked variables or events. The resulting paths $\mathcal{C}_{\text{graph}}$ expose mediators, confounders, and downstream effects aligned with the query’s underlying causal semantics.

\textbf{Unified Knowledge Set.} 
We denote the final retrieved knowledge as $\mathcal{K} = \mathcal{T}_{\text{sem}} \cup \mathcal{C}_{\text{graph}}$. This hybrid set blends semantic relevance with causal coherence, enabling downstream modules to generate grounded and causally faithful responses.

\subsection{Response Generation and Causal Graph Check}

CDF-RAG generates a response $\hat{y}$ conditioned on $\mathcal{K}$ using a language model. To ensure generated content remains faithful to causal principles, we implement a \textbf{Causal Graph Check}, verifying whether retrieved evidence supports the generated causal claims. The verification process computes a causal consistency score:
\[
S_{\text{causal}} = \frac{1}{|\mathcal{C}_{\text{graph}}|} \sum_{(C, E) \in \mathcal{C}_{\text{graph}}} \mathbb{I}(C \rightarrow E \models \hat{y}),
\]
where $\mathbb{I}$ is an indicator function that checks if the causal relation is maintained in the generated response.

If $S_{\text{causal}} < \tau$, where $\tau$ is a predefined threshold, the system triggers \textbf{Fallback Generation}, prompting the LLM to regenerate $\hat{y}$ under stricter grounding constraints:
\[
\hat{y}' = \arg\max_{y} P(y \mid \mathcal{K}, \text{strict constraints}).
\]
This ensures that the final response aligns with retrieved causal knowledge, reducing inconsistencies and hallucinations.

\subsection{Hallucination Detection and Correction}

CDF-RAG detects hallucinations by evaluating the logical consistency between $\hat{y}$ and $\mathcal{K}$. A hallucination score is computed as:
\[
S_{\text{hallucination}} = 1 - \frac{|\mathcal{K} \cap \mathcal{Y}|}{|\mathcal{Y}|},
\]
where $\mathcal{Y}$ represents extracted claims from $\hat{y}$ and $\mathcal{K}$ represents retrieved knowledge. If $S_{\text{hallucination}} > \delta$, where $\delta$ is a predefined threshold, the system applies knowledge rewriting:
\[
\hat{y}'' = \arg\max_{y} P(y \mid \mathcal{K}, \text{rewriting constraints}).
\]
This correction mechanism ensures that causal consistency is enforced without altering the base LLM capabilities, preserving factual correctness in generated responses.

\section{Experiments}
\label{sec:exp}

\subsection{Evaluation Tasks}
We evaluate the effectiveness of \textbf{CDF-RAG} across both single-hop and multi-hop question answering (QA) tasks that require varying levels of causal reasoning and knowledge integration. Our evaluations span four benchmark datasets: \textbf{CosmosQA}~\cite{huang2019cosmos}, \textbf{MedQA}~\cite{jin2020disease}, \textbf{MedMCQA}~\cite{pmlr-v174-pal22a}, and AdversarialQA~\cite{bartolo2020beat}. CosmosQA and MedQA assess commonsense and domain-specific causal reasoning, while MedMCQA and AdversarialQA test multi-hop and cross-document reasoning.

\subsection{Baselines}
We compare \textbf{CDF-RAG} against three categories of baselines:

\textbf{Standard RAG Methods:} These include conventional RAG pipelines using semantic retrieval (BM25~\cite{robertson2009probabilistic}/DPR~\cite{karpukhin2020dense}) without causal enhancement. We also consider \textit{Smart-RAG}~\cite{gao2024smartrag} and \textit{Causal-RAG}~\cite{wang2025causalrag} as stronger variants equipped with heuristic multi-hop capabilities and causal priors.

\textbf{Refined Query Methods:} We compare to \textit{Gym-RAG}~\cite{xiong2025raggym} and \textit{RQ-RAG}~\cite{chan2024rqrags}, which leverage query refinement strategies to improve retrieval quality. These serve as important baselines for assessing our reinforcement-based causal query rewrites.

\textbf{Graph-Augmented Models (G-LLMs):} We compare against a recent graph-augmented LLM framework~\cite{luo2025causal} that integrates causal filtering and chain-of-thought–driven retrieval over large knowledge graphs. This method, known as Causal-First Graph RAG, prioritizes cause-effect relationships and dynamically aligns retrieval with intermediate reasoning steps, improving interpretability and accuracy on complex medical QA tasks.

All methods are evaluated under consistent retriever and generation configurations for fair comparison. We report results using multiple LLM backbones—including GPT-4~\cite{openai2023gpt4}, LLaMA 3-8B~\cite{touvron2023llama}, Mistral~\cite{jiang2023mistral7b}, and Flan-T5~\cite{chung2024scaling}—to demonstrate model-agnostic improvements. GPT-4 is accessed via the OpenAI API, while the remaining models are fine-tuned on our curated multi-task dataset using the same training and decoding parameters to ensure alignment in evaluation settings. Additional details on the experimental setup are provided in Appendix~\ref{app:exp}.

\subsection{Metrics}

We evaluate CDF-RAG using both standard answer quality metrics and specialized measures tailored to causal reasoning and retrieval. Classical QA metrics such as accuracy, precision, recall, and F1 score assess the correctness of the final answer. To complement them, we include Context Relevance, which quantifies the semantic alignment between the user query and the retrieved content using average cosine similarity between Sentence-BERT embeddings~\cite{reimers2019sentence}. This reflects how lexically and topically well-aligned the retrieved evidence is with the original question. To assess the causal robustness of the retrieval process, we report Causal Retrieval Coverage (CRC).  CRC reflects the system’s ability to prioritize cause-effect evidence over loosely related or semantically correlated content, and serves as a proxy for the quality of causal grounding in the retrieval phase. Finally, we report Groundedness, which evaluates whether the generated answer is explicitly supported by the retrieved content.  This metric reflects factual consistency and plays a critical role in identifying hallucination-prone behaviors.

Additional metrics and results used in our study—are reported in Appendix ~\ref{sec:add_results} and provide further insight into the causal reasoning depth, performance, and robustness of the pipeline.

\section{Results and Analysis}

This section presents a detailed empirical evaluation of CDF-RAG across four benchmark QA datasets and multiple language model backbones. We compare its performance against existing RAG baselines using standard QA metrics as well as causal and contextual metrics that reflect reasoning depth, evidence grounding, and factual reliability.

\subsection{Accuracy Performance}

We report accuracy results in Table~\ref{tab:accuracy_summary} across four benchmark QA datasets—\textit{CosmosQA}~\cite{huang2019cosmos}, \textit{AdversarialQA}~\cite{bartolo2020beat}, \textit{MedQA}~\cite{jin2020disease}, and \textit{MedMCQA}~\cite{pmlr-v174-pal22a}—evaluated on four language model backbones: GPT-4~\cite{openai2023gpt4}, LLaMA 3-8B~\cite{touvron2023llama}, Mistral~\cite{jiang2023mistral7b}, and Flan-T5~\cite{chung2024scaling}. Accuracy serves as a fundamental metric for determining whether generated responses match ground-truth answers. This is particularly important in biomedical domains, where factual correctness can directly impact decision-making.

Across all datasets and models, CDF-RAG achieves the highest accuracy scores, demonstrating its generalizability across reasoning types (commonsense, adversarial, and biomedical) and model scales. On MedMCQA, for example, CDF-RAG attains 0.94 accuracy with GPT-4, and 0.90 with LLaMA 3-8B, outperforming the strongest baseline, Gym-RAG, by 16\% and 13\% respectively. Similar gains are seen across CosmosQA and AdversarialQA, highlighting CDF-RAG’s robustness in both open-domain and medically grounded QA tasks.

CDF-RAG’s improvements can be attributed to its carefully integrated architecture. First, high-quality causal pairs are extracted and validated using a GPT-4 assisted pipeline and stored in a Neo4j graph, enabling directionally-aware, multi-hop retrieval. Unlike semantic retrievers that focus on surface-level similarity, the graph captures deeper cause-effect dependencies that are essential for explanatory reasoning.

\begin{table}[ht]
\centering
\scriptsize
\caption{
\textbf{Accuracy Scores} of various RAG methods across datasets and LLM backbones.
}
\resizebox{\linewidth}{!}{%
\begin{tabular}{llcccc}
\toprule
\textbf{Dataset} & \textbf{Method} & \textbf{GPT-4} & \textbf{LLaMA 3-8B} & \textbf{Mistral} & \textbf{Flan-T5} \\
\midrule

\multirow{6}{*}{AdversarialQA}
& \cellcolor{green!20} CDF-RAG     & \cellcolor{green!20} 0.89 & \cellcolor{green!20} 0.83 & \cellcolor{green!20} 0.81 & \cellcolor{green!20} 0.79 \\
& Gym-RAG     & 0.78 & 0.75 & 0.73 & 0.70 \\
& RQ-RAG      & 0.76 & 0.71 & 0.72 & 0.66 \\
& Smart-RAG   & 0.74 & 0.73 & 0.70 & 0.64 \\
& Causal RAG  & 0.71 & 0.71 & 0.66 & 0.62 \\
& G-LLMs      & 0.68 & 0.68 & 0.65 & 0.60 \\
\midrule

\multirow{6}{*}{CosmosQA}
& \cellcolor{green!20} CDF-RAG     & \cellcolor{green!20} 0.89 & \cellcolor{green!20} 0.88 & \cellcolor{green!20} 0.85 & \cellcolor{green!20} 0.84 \\
& Gym-RAG     & 0.82 & 0.80 & 0.75 & 0.73 \\
& RQ-RAG      & 0.80 & 0.79 & 0.74 & 0.72 \\
& Smart-RAG   & 0.78 & 0.77 & 0.72 & 0.70 \\
& Causal RAG  & 0.76 & 0.75 & 0.70 & 0.68 \\
& G-LLMs      & 0.73 & 0.72 & 0.68 & 0.66 \\
\midrule

\multirow{6}{*}{MedQA}
& \cellcolor{green!20} CDF-RAG     & \cellcolor{green!20} 0.92 & \cellcolor{green!20} 0.89 & \cellcolor{green!20} 0.88 & \cellcolor{green!20} 0.84 \\
& Gym-RAG     & 0.83 & 0.79 & 0.78 & 0.73 \\
& RQ-RAG      & 0.82 & 0.78 & 0.77 & 0.72 \\
& Smart-RAG   & 0.81 & 0.77 & 0.76 & 0.71 \\
& Causal RAG  & 0.79 & 0.75 & 0.74 & 0.69 \\
& G-LLMs      & 0.76 & 0.72 & 0.71 & 0.67 \\
\midrule

\multirow{6}{*}{MedMCQA}
& \cellcolor{green!20} CDF-RAG     & \cellcolor{green!20} 0.94 & \cellcolor{green!20} 0.90 & \cellcolor{green!20} 0.88 & \cellcolor{green!20} 0.85 \\
& Gym-RAG     & 0.78 & 0.77 & 0.76 & 0.72 \\
& RQ-RAG      & 0.76 & 0.75 & 0.74 & 0.70 \\
& Smart-RAG   & 0.74 & 0.73 & 0.72 & 0.68 \\
& Causal RAG  & 0.72 & 0.71 & 0.70 & 0.66 \\
& G-LLMs      & 0.68 & 0.68 & 0.66 & 0.63 \\
\bottomrule
\end{tabular}
}
\label{tab:accuracy_summary}
\end{table}

Second, query refinement is performed via a PPO-trained RL agent, which selects between decomposition and expansion strategies. This refinement aligns the query structure with latent causal chains in the graph and vector database. Importantly, all models except GPT-4 are multi-task instruction fine-tuned on a carefully curated dataset encompassing decomposition, simplification, and expansion tasks (or any combinations in a feedback loop)—enabling consistent, controllable query rewriting across backbones.

Compared to other methods, CDF-RAG offers a more coherent and complete reasoning stack. Gym-RAG performs well due to its reward-guided trajectory optimization, but it lacks causal grounding in its retrieval process and does not validate final outputs, leading to gaps in factual correctness. RQ-RAG uses static rule-based query rewriting, which improves retrieval over raw queries, but it is not adaptive and does not distinguish between causal and associative evidence, limiting its effectiveness on multi-hop queries.

Smart-RAG includes an RL policy to coordinate retrieval and generation steps. However, it lacks access to causal graphs and performs no verification of output consistency. Its reliance on semantic retrieval alone results in shallow, often incomplete, reasoning. Causal RAG, while integrating causal paths, depends on weak summarization-based extraction methods, leading to noisy graph construction. It does not dynamically refine queries or filter hallucinations. G-LLMs use static knowledge graphs but are not designed to support causal retrieval or adaptive refinement, resulting in the lowest accuracy across all configurations.

In contrast, CDF-RAG’s feedback loop between query refinement, causal retrieval, and answer verification ensures end-to-end causal alignment. This alignment not only improves retrieval coverage but also helps the model generate answers that are more accurate and grounded in factually valid causal pathways.

These results emphasize the need for RAG frameworks to go beyond surface-level semantic retrieval and incorporate causal structure, dynamic reasoning strategies, and output validation. CDF-RAG embodies these principles, resulting in substantial improvements in accuracy across datasets and model architectures.

\subsection{Retrieval and Contextual Performance}

\renewcommand{\tabcolsep}{3.5pt}
\renewcommand{\arraystretch}{0.95}

\begin{table*}[t]
\centering
\scriptsize
\caption{
\textbf{Retrieval and Contextual Metrics} of \textbf{CDF-RAG} across models and methods. \\
CRC = Causal Retrieval Coverage, Context = Context Relevance.
}
\label{tab:contextual-metrics}
\begin{tabular}{l@{\hskip 3pt}l@{\hskip 3pt}c@{\hskip 3pt}c@{\hskip 3pt}c@{\hskip 3pt}c@{\hskip 3pt}c@{\hskip 3pt}c}
\toprule
\textbf{Dataset} & \textbf{Model} & \textbf{CDF-RAG} & \textbf{Gym-RAG} & \textbf{RQ-RAG} & \textbf{Smart-RAG} & \textbf{Causal RAG} & \textbf{G-LLMs} \\
\midrule
\multicolumn{8}{c}{\textbf{CRC}} \\
\midrule
\multirow{4}{*}{AdversarialQA}
& GPT-4      & \cellcolor{green!20} 0.89 & 0.80 & 0.77 & 0.74 & 0.72 & 0.68 \\
& LLaMA 3-8B & \cellcolor{green!20} 0.85 & 0.76 & 0.73 & 0.70 & 0.68 & 0.64 \\
& Mistral    & \cellcolor{green!20} 0.82 & 0.74 & 0.71 & 0.68 & 0.66 & 0.61 \\
& Flan-T5    & \cellcolor{green!20} 0.78 & 0.70 & 0.67 & 0.64 & 0.62 & 0.59 \\
\midrule
\multirow{4}{*}{CosmosQA}
& GPT-4      & \cellcolor{green!20} 0.91 & 0.81 & 0.79 & 0.77 & 0.74 & 0.71 \\
& LLaMA 3-8B & \cellcolor{green!20} 0.87 & 0.78 & 0.76 & 0.74 & 0.71 & 0.68 \\
& Mistral    & \cellcolor{green!20} 0.86 & 0.79 & 0.77 & 0.75 & 0.72 & 0.69 \\
& Flan-T5    & \cellcolor{green!20} 0.82 & 0.75 & 0.73 & 0.71 & 0.69 & 0.66 \\
\midrule
\multirow{4}{*}{MedMCQA}
& GPT-4      & \cellcolor{green!20} 1.00 & 0.93 & 0.90 & 0.88 & 0.85 & 0.82 \\
& LLaMA 3-8B & \cellcolor{green!20} 0.98 & 0.89 & 0.86 & 0.84 & 0.81 & 0.78 \\
& Mistral    & \cellcolor{green!20} 0.96 & 0.91 & 0.88 & 0.86 & 0.83 & 0.80 \\
& Flan-T5    & \cellcolor{green!20} 0.95 & 0.87 & 0.84 & 0.82 & 0.79 & 0.76 \\
\midrule
\multicolumn{8}{c}{\textbf{Context (Context Relevance)}} \\
\midrule
\multirow{4}{*}{AdversarialQA}
& GPT-4      & \cellcolor{green!20} 0.76 & 0.67 & 0.64 & 0.62 & 0.60 & 0.56 \\
& LLaMA 3-8B & \cellcolor{green!20} 0.73 & 0.65 & 0.63 & 0.60 & 0.58 & 0.54 \\
& Mistral    & \cellcolor{green!20} 0.70 & 0.63 & 0.60 & 0.58 & 0.56 & 0.52 \\
& Flan-T5    & \cellcolor{green!20} 0.68 & 0.60 & 0.58 & 0.56 & 0.53 & 0.50 \\
\midrule
\multirow{4}{*}{CosmosQA}
& GPT-4      & \cellcolor{green!20} 0.78 & 0.69 & 0.67 & 0.66 & 0.63 & 0.60 \\
& LLaMA 3-8B & \cellcolor{green!20} 0.75 & 0.67 & 0.65 & 0.63 & 0.61 & 0.58 \\
& Mistral    & \cellcolor{green!20} 0.74 & 0.67 & 0.65 & 0.64 & 0.62 & 0.59 \\
& Flan-T5    & \cellcolor{green!20} 0.72 & 0.64 & 0.62 & 0.61 & 0.59 & 0.56 \\
\midrule
\multirow{4}{*}{MedMCQA}
& GPT-4      & \cellcolor{green!20} 0.64 & 0.60 & 0.58 & 0.56 & 0.54 & 0.51 \\
& LLaMA 3-8B & \cellcolor{green!20} 0.62 & 0.58 & 0.56 & 0.54 & 0.52 & 0.49 \\
& Mistral    & \cellcolor{green!20} 0.63 & 0.59 & 0.57 & 0.55 & 0.53 & 0.50 \\
& Flan-T5    & \cellcolor{green!20} 0.61 & 0.57 & 0.55 & 0.53 & 0.50 & 0.47 \\
\bottomrule
\end{tabular}
\end{table*}

To evaluate the quality of retrieval and its downstream impact on answer faithfulness, we report three upstream metrics: CRC, Context Relevance, and Groundedness. Together, these metrics assess how well the system identifies causally relevant content, aligns it semantically with the user query, and generates factually supported answers.

CRC measures the proportion of retrieved elements—including causal triples from the Neo4j graph and unstructured passages from the vector database—that belong to a verified causal path aligned with the query. For each query, CRC is computed by checking whether the retrieved items match entries in a gold-standard causal graph constructed using GPT-4 verification. This metric reflects the system’s ability to prioritize directional, explanatory content over semantically correlated but causally irrelevant material.

Context Relevance captures the semantic alignment between the user query and the retrieved content. We compute this by encoding both the query and the top-k retrieved items using Sentence-BERT embeddings~\cite{reimers2019sentence}, followed by averaging cosine similarity scores across the retrieved set. While CRC emphasizes structural fidelity, Context Relevance ensures that the retrieved material is lexically and topically close to the query, which is particularly useful in guiding generation during early inference steps.

Groundedness evaluates the factual consistency between the generated answer and the retrieved evidence. It assesses whether the key claims made in the answer are explicitly supported by the retrieved content, ensuring that the response is not only fluent but also verifiable. To compute this metric, we apply a span-level alignment approach that checks whether the answer content can be traced back to specific supporting phrases or structures within the retrieved passages or causal triples. Groundedness is crucial because a model may retrieve high-quality context yet still introduce hallucinations or unsupported causal links during generation. This metric reflects the degree to which the model faithfully uses its retrieved inputs, serving as a proxy for factual reliability and evidence-grounded reasoning.

As shown in Table~\ref{tab:contextual-metrics}, CDF-RAG consistently achieves the highest CRC and Context Relevance across all datasets and LLMs. For example, on AdversarialQA with GPT-4, CDF-RAG attains a CRC of 0.89 and Context score of 0.76, while Gym-RAG and RQ-RAG score 0.80/0.67 and 0.77/0.64, respectively. On MedMCQA, CDF-RAG achieves perfect causal coverage (CRC = 1.00) and the highest semantic alignment (Context = 0.64). These metrics explain its ability to retrieve both relevant and causally grounded information.

\begin{figure}[ht]
    \centering
    \includegraphics[width=0.85\linewidth]{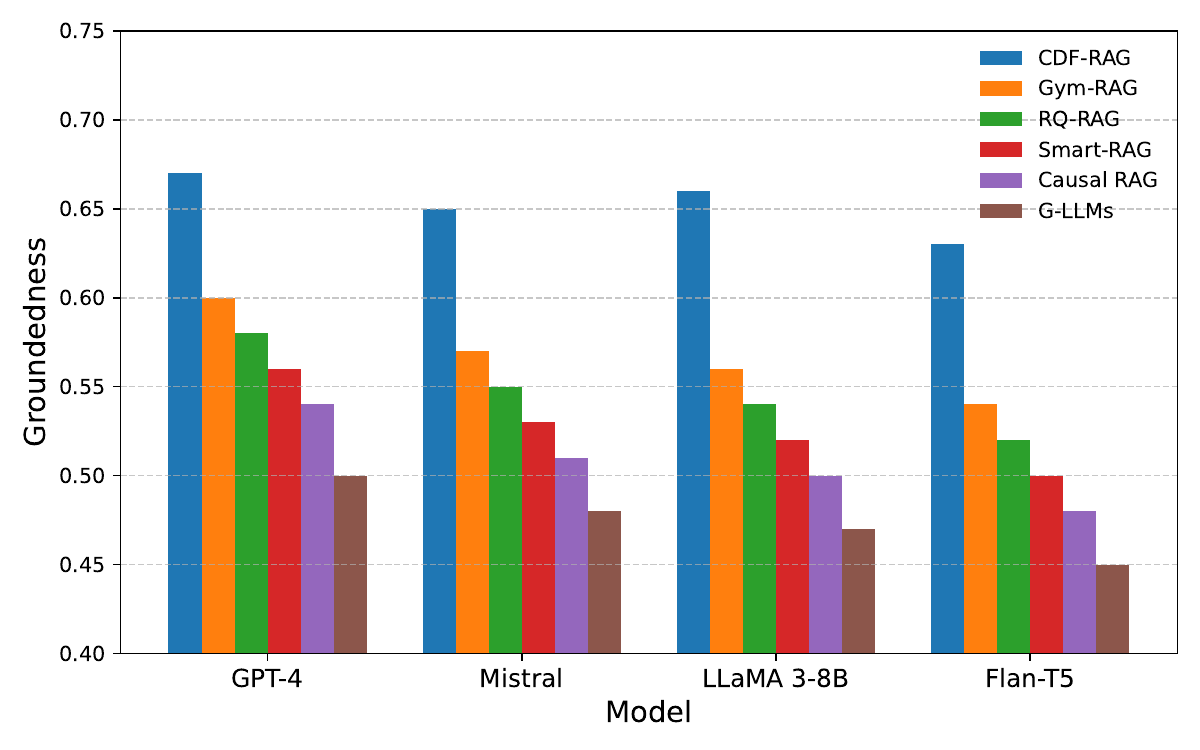}
    \caption{Groundedness comparison of different methods across four LLMs on the MedQA dataset.}
    \label{fig:groundedness_comparison}
\end{figure}

Figure~\ref{fig:groundedness_comparison} presents the Groundedness comparison across four LLMs on the MedQA dataset. CDF-RAG outperforms all baselines across every model backbone, achieving a groundedness score of 0.67–0.65, depending on the LLM. The improvement is especially notable with GPT-4 and LLaMA 3-8B, where the margin over Gym-RAG and Smart-RAG is over 7\%. This gain highlights the benefit of our hallucination-aware verification loop and causally coherent retrieval. By integrating RL-refined queries, causal graph traversal, and structured rewriting, CDF-RAG ensures that generation remains closely tied to verifiable, context-supported content. In contrast, methods such as G-LLMs and Causal-RAG either lack semantic adaptation or perform shallow causal reasoning, resulting in lower groundedness and higher susceptibility to hallucinations. Together, these results confirm that CDF-RAG’s retrieval pipeline is both structurally precise and semantically aligned, leading to answers that are not only accurate but also causally and contextually grounded.

\begin{figure*}[th]
\centering
\includegraphics[width=0.79\linewidth]{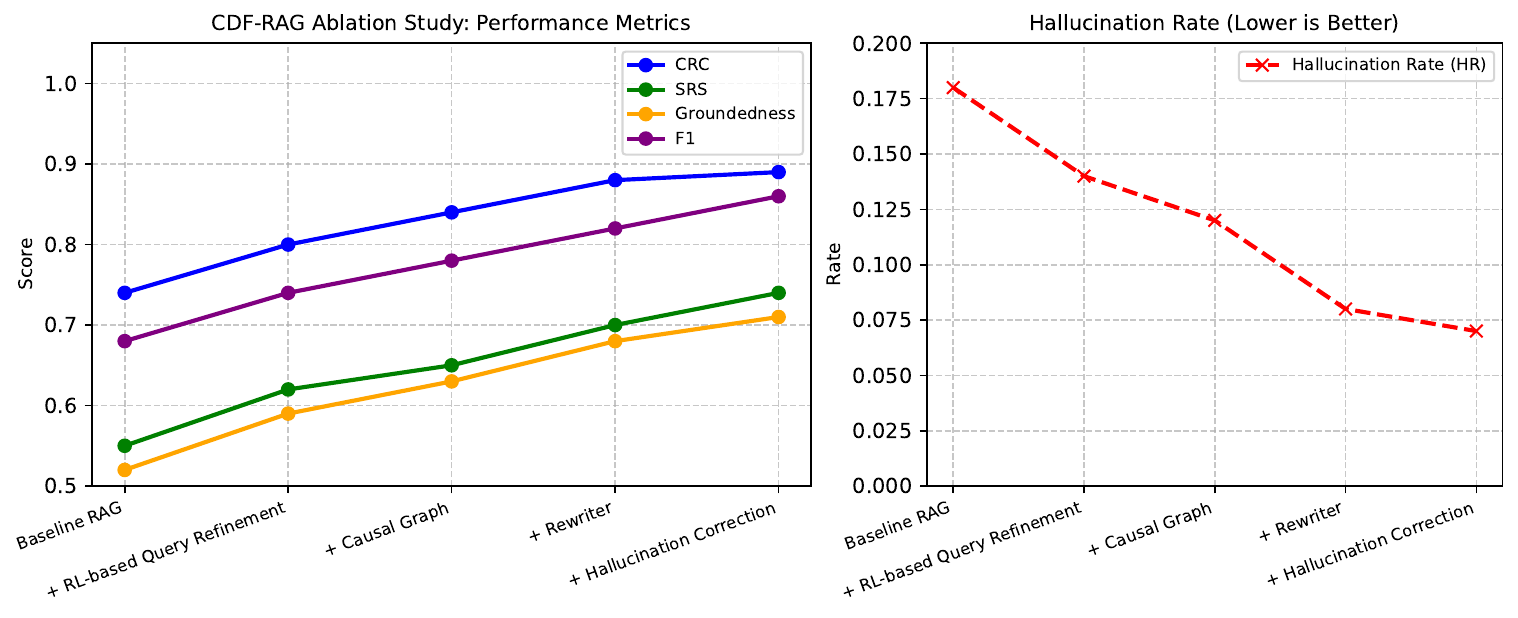}
\caption{
Ablation study of CDF-RAG across incremental stages. Left: performance metrics including CRC, SRS, groundedness, and F1 score. Right: HR, where lower values indicate greater factual consistency.
}
\label{fig:cdf_ablation}
\end{figure*}


\subsection{Ablation Study}

To evaluate the contribution of each component in the CDF-RAG framework, we conduct a stepwise ablation study by incrementally enabling key modules. We begin with a baseline RAG setup that uses semantic vector retrieval via MiniLM, followed by LLM-based generation, without incorporating query refinement or structural retrieval mechanisms. We then progressively add RL-based query refinement, causal graph retrieval, structured knowledge rewriting, and hallucination correction. Notably, query refinement is only triggered when the RL agent is enabled, and no static prompt engineering is applied at any stage.

Each configuration is evaluated on six metrics: CRC, causal chain depth (CCD), semantic refinement score (SRS), groundedness, hallucination rate (HR), and F1. 
CCD measures the average number of directed hops in retrieved causal paths from the Neo4j graph. 
SRS is computed as the cosine similarity between the original and refined queries, quantifying semantic alignment. 
Groundedness reflects the coherence between retrieved knowledge and generated responses, using sentence embedding similarity. 
HR denotes the percentage of responses flagged as hallucinated by the LLM verifier.

F1 captures the balance of precision and recall based on overlap with reference answers. As shown in Table~\ref{tab:ablation_cdf_rag}, each added component improves overall system performance. For example, enabling the causal graph module increases CCD from 1.70 to 1.92 by exposing deeper multi-hop pathways, while the hallucination verifier further reduces HR to 0.07 and improves groundedness to 0.71. The final stage values match those reported in the main results table, confirming that each module contributes both independently and synergistically to the robustness and reliability of CDF-RAG’s causal reasoning.

Figure~\ref{fig:cdf_ablation} presents the results of our ablation study over the CDF-RAG framework, highlighting the incremental effect of each component. As we progressively add RL-based query refinement, causal graph retrieval, structured rewriting, and hallucination correction, we observe consistent gains across core metrics such as CRC, SRS, groundedness, and F1. The HR shows a marked decline, reflecting enhanced factual reliability at each stage. These results underscore the modular design and cumulative benefit of CDF-RAG's causally grounded and agentic reasoning architecture.


\begin{table}[t]
\centering
\scriptsize
\caption{
Ablation study on the CDF-RAG framework. Each stage adds a core module, demonstrating consistent gains across CRC, CCD, SRS, groundedness, and F1 score. HR reflects improved factual reliability. 
}
\label{tab:ablation_cdf_rag}
\resizebox{\linewidth}{!}{%
\begin{tabular}{lcccccc}
\toprule
\textbf{Ablation Stage} & \textbf{CRC} & \textbf{CCD} & \textbf{SRS} & \textbf{Groundedness} & \textbf{HR} & \textbf{F1} \\
\midrule
Baseline RAG & 0.74 & 1.50 & 0.55 & 0.52 & 0.18 & 0.68 \\
\makecell[l]{+ RL-based Query\\Refinement} & 0.80 & 1.70 & 0.62 & 0.59 & 0.14 & 0.74 \\
+ Causal Graph & 0.84 & 1.92 & 0.65 & 0.63 & 0.12 & 0.78 \\
+ Rewriter & 0.88 & 2.00 & 0.70 & 0.68 & 0.08 & 0.82 \\
\rowcolor{green!20} \makecell[l]{+ \textbf{Hallucination}\\\textbf{Correction (Ours)}} & \textbf{0.89} & \textbf{2.02} & \textbf{0.74} & \textbf{0.71} & \textbf{0.07} & \textbf{0.86} \\
\bottomrule
\end{tabular}
}
\end{table}

\section{Conclusion}
In this paper, we introduce CDF-RAG, a causality-aware RAG framework that integrates reinforcement-learned query refinement, multi-hop causal graph retrieval, and hallucination detection into a dynamic feedback loop. By aligning retrieval with causal structures and enforcing consistency in generation, CDF-RAG enhances factual accuracy and reasoning depth. Our evaluations demonstrate state-of-the-art performance across four QA benchmarks, surpassing existing RAG methods in causal correctness and reliability. These results highlight the effectiveness of structured causal reasoning for adaptive retrieval-augmented generation.

\newpage
\section*{Limitations}

While CDF-RAG demonstrates improvements in retrieval precision and response coherence through causal query refinement, several limitations remain. First, the method depends on access to structured causal graphs, which may not be readily available or complete in all domains, particularly those with sparse or noisy causal knowledge. This reliance could limit applicability in open-domain or low-resource settings. Second, the hallucination detection module employs GPT-based validation, which, despite its effectiveness, incurs significant computational overhead. This may hinder deployment in real-time or resource-constrained environments. Finally, although our reinforcement learning framework enables adaptive query refinement, its generalization to highly heterogeneous or informal queries requires further investigation. Addressing these limitations is essential for broader applicability and efficiency in practical settings.

\bibliography{ref}

\appendix
\section{Appendix - Dataset} \label{sec:appendix}

This appendix provides additional implementation and experimental details to support the results and claims presented in the main paper. It includes comprehensive documentation of our data construction process, fine-tuning setup, and evaluation procedures. We also provide prompt templates used for multi-task instruction tuning, detailed ablation metrics, and further discussions of design choices and observations.

\subsection{Data Collection and Causal Graph Construction}

Our data collection process supports two major objectives: (1) training the query refinement module with multi-task instruction examples, and (2) constructing structured causal knowledge graphs that power CDF-RAG’s graph-based retrieval. We collect and process data from four benchmark QA datasets—\textit{CosmosQA}, \textit{AdversarialQA}, \textit{MedQA}, and \textit{MedMCQA}—chosen for their coverage of commonsense, adversarial, and biomedical reasoning tasks. Each dataset is used to extract causally relevant triples and generate query refinement prompts across decomposition, expansion, and simplification modes.

To enable structured causal retrieval, we implement a dedicated preprocessing pipeline named \textbf{CausalFusion}. This component combines fine-tuned causal classification with LLM-based validation to extract high-confidence cause-effect pairs from each dataset. Specifically, we build on the UniCausal~\cite{tan2023unicausal} framework and focus on the Causal Pair Classification task. Sentences from each dataset are annotated with candidate argument spans (\texttt{<ARG0>} and \texttt{<ARG1>}), which are passed through a BERT-based encoder trained to predict whether a causal relationship exists between them. The model outputs binary judgments that filter candidate pairs down to high-quality causal candidates.

Following this step, we apply a GPT-4 refinement stage to all accepted causal pairs. GPT-4 serves as a semantic verifier and reformulator: it rephrases each pair into a fluent, logically coherent causal statement, flags inconsistencies, and rejects biologically implausible or semantically invalid pairs. The output for each instance includes the original dataset name, cause and effect variables, predicted directionality, and the refined natural language causal explanation.

All validated and rephrased causal pairs are stored as directed triples in a Neo4j knowledge graph. To support fast and semantically aware retrieval, we encode each node (cause or effect) into a 384-dimensional embedding using MiniLM-based sentence encoders. These embeddings are stored in a vector database alongside their graph identifiers, enabling hybrid semantic and path-based retrieval during inference. This graph forms the foundation for multi-hop causal reasoning in CDF-RAG and is continuously updated as new validated pairs are added.

This hybrid symbolic-neural representation ensures that retrieval can traverse explicit causal paths while remaining robust to lexical variation in user queries. It also provides a structured backbone for measuring retrieval depth, validating generation, and supporting hallucination detection via graph-based entailment.

\subsection{Causal Prompt Design for Pair Verification}

To ensure the factual and causal correctness of extracted pairs in our CDF pipeline, we design a GPT-4-based verification module using structured natural language prompts. Each extracted causal pair undergoes a validation stage, where it is converted into a prompt and sent to GPT-4 for semantic and causal assessment. The goal is to ensure that only high-confidence, directionally accurate, and domain-valid causal links are retained for inclusion in the Neo4j causal graph.

We adopt a contextualized causal prompting strategy inspired by the causal wrapper component in ALCM~\cite{khatibi2024alcm}. Each prompt includes:
\begin{itemize}
    \item \textbf{Instruction}—Defining GPT-4’s role in assessing the causal pair.
    \item \textbf{Contextual Metadata}—Information about the dataset, domain, and source extraction model.
    \item \textbf{Causal Pair}—The specific cause-effect relationship being assessed.
    \item \textbf{Task Definition}—Explicit questions about the validity, direction, and justification of the causal link.
    \item \textbf{Output Format}—A structured template including a binary correctness flag, refined causal direction, confidence score, and explanation.
\end{itemize}

This causal prompt design enables the LLM to reason explicitly about the plausibility and correctness of each candidate link. It also facilitates standardized post-processing by producing consistent, machine-readable outputs. Verified causal pairs are then re-integrated into the graph database, ensuring that downstream query refinement and multi-hop reasoning are grounded in trustworthy knowledge.

An illustrative example of such a prompt is shown below:

\begin{quote}
\texttt{Causal Pair to Verify: \{Cause: High blood pressure, Effect: Stroke\}} \\
\texttt{Correctness: True} \\
\texttt{Refined Causal Statement: "High blood pressure causes stroke"} \\
\texttt{Confidence: High} \\
\texttt{Explanation: Chronic hypertension is a well-known risk factor for stroke based on medical literature.}
\end{quote}

\begin{tcolorbox}[colback=white, colframe=black, sharp corners, boxrule=0.8pt, width=\linewidth, arc=2mm, title=Causal Verification Prompt Template]

You are an expert in \texttt{\{DOMAIN\}} with deep knowledge of causal relationships and evidence-based reasoning.

You are given a candidate causal relationship extracted from a document or causal discovery algorithm.

Your task is to evaluate whether the following causal relationship is factually and logically correct based on your internal knowledge and reasoning. You may accept, reject, revise, or reorient the pair. Use step-by-step reasoning to justify your answer.

\medskip
\textbf{Contextual Metadata:} \\
• Domain: \texttt{\{DOMAIN\}} \\
• Dataset: \texttt{\{DATASET NAME\}} \\
• Source Model: \texttt{\{MODEL or EXTRACTION METHOD\}}

\medskip
\textbf{Causal Pair to Verify:} \\
• Cause: \texttt{\{ARG0\}} \\
• Effect: \texttt{\{ARG1\}}

\medskip
\textbf{Task:} \\
1. Is the causal relationship valid and supported? (Answer: True/False) \\
2. If the direction is incorrect, provide the corrected direction. \\
3. Provide a one-sentence explanation justifying your decision. \\
4. Estimate your confidence in the answer (High / Medium / Low)

\medskip
\textbf{Output Format:} \\
Correctness: \texttt{\{True / False\}} \\
Refined Causal Statement: \texttt{"\{ARG0\}" causes "\{ARG1\}"} or \texttt{"\{ARG1\}" causes "\{ARG0\}"} \\
Confidence: \texttt{\{High / Medium / Low\}} \\
Explanation: \texttt{\{Short justification grounded in domain knowledge\}}

\end{tcolorbox}

\subsection{Reinforcement Learning for Query Refinement.}
To dynamically optimize query rewriting strategies in CDF-RAG, we train a RL agent using the Proximal Policy Optimization (PPO) algorithm. The agent learns a policy $\pi(a|s)$ that maps the semantic embedding of a raw query $s$ to one of three refinement actions: \textit{Expand}, \textit{Simplify}, or \textit{Decompose}. Each action corresponds to a rewriting strategy aimed at improving causal specificity and retrievability. The agent interacts with a custom Gym environment, where each state $s$ is a 384-dimensional embedding of the input query (from MiniLM), and the action space is discrete over refinement types.

The reward function integrates downstream performance metrics critical for causal reasoning. 
After each refinement action, the system executes the retrieval and generation pipeline and computes four normalized metrics: 
retrieval relevance ($r$), causal depth ($d$), semantic similarity ($s$), and hallucination rate ($h$). 
The reward function is defined as:
\[
\mathcal{R} = \lambda_1 r + \lambda_2 d + \lambda_3 s + \lambda_4 (1 - h)
\]


where each component is normalized to the range $[0,1]$, and $\lambda_i$ are tunable weights controlling the importance of each term. Relevance measures whether the refinement improves the match between retrieved context and query intent; causal depth quantifies the number of multi-hop causal links retrieved; semantic similarity evaluates alignment with the original query; and hallucination penalizes factual inconsistency in generated outputs.

We train the agent using PPO with a two-layer MLP policy network (hidden size 256), batch size 64, learning rate $3 \times 10^{-4}$, and entropy regularization of 0.01. Training is run for 100 epochs with 500 steps per query. The training curriculum covers diverse domains by sampling queries from MedQA, CosmosQA, and AdversarialQA. All models except GPT-4 are trained using this RL framework after multi-task instruction fine-tuning.

At inference time, the trained policy $\pi(a|s)$ selects the optimal refinement action given an unseen input query. This enables the system to adaptively reformulate questions in a way that aligns with both the causal structure of the knowledge graph and the semantic requirements of the task, thereby improving downstream accuracy, coherence, and explainability.

\subsection{Prompt Design for Multi-task Instruction Fine-tuning}

To enable the query refinement module in CDF-RAG to adaptively rewrite input questions, we construct a multi-task instruction dataset covering three core refinement actions: \textit{Simplify}, \textit{Decompose}, and \textit{Expand}. These refinement strategies correspond to key capabilities required for causal reasoning: clarifying ambiguous questions, breaking down complex ones into causal subcomponents, and enriching underspecified queries with relevant scope. For each action type, we design a specialized prompt template to guide GPT-4 in generating high-quality supervision examples. These templates are used to fine-tune the LLMs (LLaMA 3-8B, Mistral, and Flan-T5) using LoRA, while GPT-4 is accessed via API at inference time without fine-tuning.

\paragraph{Simplification Prompt} 
As shown in Prompt Box~\ref{box:simplify-prompt}, we provide GPT-4 with detailed instructions for simplifying complex questions while preserving their original intent.
This template is used to rephrase complex, ambiguous, or overly verbose queries into concise and direct questions while preserving their original intent. The goal is to strip away unnecessary syntactic or semantic complexity to improve retrievability and alignment with the knowledge base. The model is instructed to output a single-line question that is self-contained and interpretable, which is essential for enhancing the precision of retrieval in high-noise or cross-domain settings.

\begin{tcolorbox}[colback=white, colframe=black, sharp corners, boxrule=0.8pt, width=\linewidth, arc=2mm, title=Simplification Prompt Template]
\label{box:simplify-prompt}

Your task is to simplify complex or ambiguous questions into a clearer, more
direct version that preserves the original intent. This should help reduce
unnecessary complexity while keeping the meaning intact.

\medskip
\textbf{Please follow the steps below carefully:}
\begin{enumerate}
    \item Identify any ambiguity, compound phrasing, or indirect constructs in the input question.
    \item Reformulate the question as a concise, direct, and self-contained single question.
    \item Ensure that the simplified version can be interpreted and answered independently.
\end{enumerate}

\textbf{Guidelines:}
\begin{itemize}
    \item Use precise language that avoids unnecessary technical or abstract phrasing.
    \item Do not generate multiple sub-questions.
    \item Keep the simplified question to a single line of text.
    \item Preserve the core meaning of the original question.
\end{itemize}

\textbf{Here is your task:}
\begin{itemize}
    \item Provided Contexts: \texttt{\{OPTIONAL — leave blank or include background passages\}}
    \item Original Question: \texttt{\{INSERT COMPLEX OR AMBIGUOUS QUESTION\}}
    \item Simplified Query:
\end{itemize}

\end{tcolorbox}


\paragraph{Prompting Strategy.} 
To enable query simplification within CDF-RAG, we adopt a dual-prompting approach tailored for both system implementation and interpretability. For fine-tuning and inference, we use a concise instruction-tuning format (\texttt{"Refine the following query for better causal retrieval"}) to streamline training across hundreds of examples. To complement this, we define a structured prompt template with explicit steps and guidelines for simplification, which is used in our paper to illustrate the design intent behind simplification behavior. This alignment between lightweight instructional prompts and a principled template ensures both efficiency and transparency in how simplification is operationalized within the framework.

\begin{tcolorbox}[colback=white, colframe=black, sharp corners, boxrule=0.8pt, width=\linewidth, arc=2mm, title=Simplification Prompt Template]
\label{box:simplify-prompt}

Your task is to simplify complex or ambiguous questions into a clearer, more
direct version that preserves the original intent. This should help reduce
unnecessary complexity while keeping the meaning intact.

\medskip
\textbf{Please follow the steps below carefully:}
\begin{enumerate}
    \item Identify any ambiguity, compound phrasing, or indirect constructs in the input question.
    \item Reformulate the question as a concise, direct, and self-contained single question.
    \item Ensure that the simplified version can be interpreted and answered independently.
\end{enumerate}

\textbf{Guidelines:}
\begin{itemize}
    \item Use precise language that avoids unnecessary technical or abstract phrasing.
    \item Do not generate multiple sub-questions.
    \item Keep the simplified question to a single line of text.
    \item Preserve the core meaning of the original question.
\end{itemize}

\textbf{Here is your task:}
\begin{itemize}
    \item Provided Contexts: \texttt{Medical QA task related to diabetic nephropathy}
    \item Original Question: \texttt{Why does diabetes cause kidney damage in elderly patients, and what factors contribute to this progression over time?}
    \item Simplified Query: \texttt{How does diabetes cause kidney damage in elderly patients?}
\end{itemize}

\end{tcolorbox}

\paragraph{Decomposition Prompt}   
The decomposition prompt (see Prompt Box~\ref{box:decompose-prompt}) teaches the model to break down multihop or causally entangled questions into 2–4 atomic sub-questions that collectively reconstruct the original reasoning chain. Each sub-question should be answerable independently and follow a logical progression that mirrors multi-hop causal inference. This prompt is particularly important for enabling causal retrieval over multi-node paths in the Neo4j graph and for promoting modular reasoning within the generation phase.

\begin{tcolorbox}[
  colback=white,
  colframe=black,
  sharp corners,
  boxrule=0.8pt,
  arc=2mm,
  title=Decomposition Prompt Template,
  label={box:decompose-prompt}
]

Your task is to decompose complex, multi-hop questions into simpler, manageable
sub-questions. These decomposed queries should help isolate and uncover causal
or explanatory mechanisms relevant to the original question.

\medskip
\textbf{Please follow the steps below carefully:}
\begin{enumerate}
    \item Analyze the multihop question to identify its underlying causal or semantic components.
    \item Reformulate the question into a list of 2–4 clear, concise, self-contained sub-questions that can be independently answered.
    \item Maintain logical flow between sub-questions (i.e., each one should build toward answering the original question).
\end{enumerate}

\textbf{Guidelines:}
\begin{itemize}
    \item Avoid repeating the same phrasing across sub-questions.
    \item Each sub-question should be answerable on its own.
    \item Use one line per sub-question, and insert a line break between each.
    \item Do not include numbered bullets or explanations—only the raw list of sub-questions.
\end{itemize}

\textbf{Here is your task:}
\begin{itemize}
    \item Provided Contexts: \texttt{\{OPTIONAL — leave blank or include background passages\}}
    \item Multihop Question: \texttt{\{INSERT MAIN QUESTION\}}
    \item Decomposed Queries:
\end{itemize}

\end{tcolorbox}


\paragraph{Prompting Strategy.}
For decomposition, we employ a structured prompt that guides the model to break down complex, multihop questions into logically ordered sub-questions (see Prompt Box~\ref{box:decompose-prompt}). While this instructional format is used for transparency and design illustration, the deployed system leverages a compact instruction-tuning variant during fine-tuning and inference (e.g., \texttt{"Break this question into sub-questions for causal reasoning."}). This alignment allows us to retain explainability in prompt engineering while maintaining efficiency and generalizability in real-time execution.

\begin{tcolorbox}[
  colback=white,
  colframe=black,
  sharp corners,
  boxrule=0.8pt,
  arc=2mm,
  title=Decomposition Prompt Template,
  label={box:decompose-prompt-example}
]

Your task is to decompose complex, multi-hop questions into simpler, manageable
sub-questions. These decomposed queries should help isolate and uncover causal
or explanatory mechanisms relevant to the original question.

\textbf{Here is your task:}
\begin{itemize}
    \item Provided Contexts: \texttt{Healthcare domain — diabetes and kidney disease}
    \item Multihop Question: \texttt{Why does diabetes lead to kidney failure in aging populations over time?}
    \item Decomposed Queries:
\end{itemize}

What physiological changes does diabetes cause in the kidneys?\\
How does chronic hyperglycemia damage kidney function over time?\\
What role does aging play in accelerating diabetic kidney complications?\\
Why are older adults more susceptible to renal decline with diabetes?

\end{tcolorbox}

\paragraph{Expansion Prompt}  
For queries that are vague or underspecified, the expansion prompt (see Prompt Box~\ref{box:expand-prompt})
 guides the model to make the question more complete by adding relevant causal factors, domain-specific constraints, or example conditions. The objective is to surface latent context or scope that may be implicitly expected but is missing in the original query. This expanded form allows the retrieval system to access a broader and more causally aligned evidence space.

\begin{tcolorbox}[
  colback=white,
  colframe=black,
  sharp corners,
  boxrule=0.8pt,
  arc=2mm,
  title=Expansion Prompt Template,
  label={box:expand-prompt}
]

Your task is to expand a vague or underspecified question into a more detailed
version that makes its intent clear and specific. This should help clarify the
scope of the question by introducing relevant dimensions, factors, or examples.

\medskip
\textbf{Please follow the steps below carefully:}
\begin{enumerate}
    \item Identify missing context or implicit assumptions in the question.
    \item Reformulate the question to explicitly mention key entities, causal mechanisms, or domains relevant to the query.
    \item Ensure the expanded question guides a more targeted and informative answer.
\end{enumerate}

\textbf{Guidelines:}
\begin{itemize}
    \item Use a single line for the expanded question.
    \item Avoid changing the core topic, but add specificity or scope.
    \item Preserve the original intent, while making the question more complete or informative.
\end{itemize}

\textbf{Here is your task:}
\begin{itemize}
    \item Provided Contexts: \texttt{\{OPTIONAL — leave blank or include background passages\}}
    \item Original Question: \texttt{\{INSERT VAGUE OR INCOMPLETE QUESTION\}}
    \item Expanded Query:
\end{itemize}

\end{tcolorbox}


\paragraph{Prompting Strategy.}
The expansion prompt is designed to elicit more informative and context-aware reformulations for vague or under-specified queries (see Prompt Box~\ref{box:expand-prompt}). While this prompt is used to train the model to surface latent causal factors and clarify scope, the system implementation uses a condensed instruction-tuned variant (e.g., \texttt{"Make the question more specific for causal reasoning"}). This dual-prompting setup ensures that the model learns how to expand queries both accurately and efficiently, while also preserving interpretability and alignment during prompt analysis and dataset curation.

\begin{tcolorbox}[
  colback=white,
  colframe=black,
  sharp corners,
  boxrule=0.8pt,
  arc=2mm,
  title=Expansion Prompt Template,
  label={box:expand-prompt-example}
]

Your task is to expand a vague or underspecified question into a more detailed
version that makes its intent clear and specific. This should help clarify the
scope of the question by introducing relevant dimensions, factors, or examples.

\textbf{Here is your task:}
\begin{itemize}
    \item Provided Contexts: \texttt{Societal health disparities and stress}
    \item Original Question: \texttt{Why is stress a public health concern?}
    \item Expanded Query:
\end{itemize}

Why is chronic stress considered a public health concern in relation to socioeconomic status, mental health, and long-term disease risk?

\end{tcolorbox}

Together, these prompt templates form the backbone of our instruction fine-tuning strategy, enabling each model to learn not only how to execute a refinement action, but also when and why such rewrites are useful for causal alignment. Each generated example is filtered for consistency and correctness before being added to the training dataset. During inference, the PPO-trained policy network selects among these three refinement actions for each input query, enabling dynamic adaptation to the structure and intent of unseen questions.

\section{Appendix - Experimental Details} ~\label{app:exp}

\subsection{Training and Fine-tuning Setup}

We fine-tune all LLM backbones (except GPT-4, which is accessed via API) using LoRA with instruction-style supervision. Each model is trained on our multi-task dataset for one epoch with a learning rate of 2e-5 and 3\% warmup steps. 

\subsection{ Comparison with Related Work}

CDF-RAG introduces a comprehensive and agentic approach to RAG by combining causal graph retrieval, RL-driven query refinement, multi-hop reasoning, and hallucination correction into a unified framework. This integrated design enables the model to explicitly reason over structured cause-effect relationships while adaptively optimizing queries and validating outputs through a closed-loop process. Unlike existing methods that focus on isolated components of the RAG pipeline, CDF-RAG emphasizes the causal alignment and coherence of both retrieved and generated content.

In contrast, methods like RQ-RAG and SmartRAG provide query refinement capabilities—via decomposition or RL—but do not incorporate causal graph retrieval or hallucination mitigation. RAG-Gym offers process-level optimization through nested MDPs and includes a hallucination-aware reward model, but lacks structural causal reasoning. Causal Graph RAG and Causal Graphs Meet Thoughts integrate causal graphs but fall short in dynamic feedback, multi-agent coordination, and hallucination control. Overall, CDF-RAG is distinguished by its holistic design that tightly couples causal retrieval, adaptive refinement, and output validation—resulting in improved factuality, reasoning depth, and consistency.



\subsection{Implementation and Agentic Design}
Our CDF-RAG framework is implemented using the LangChain library, which provides modular primitives for constructing agentic workflows in language model systems. We structure the pipeline as a multi-step LangGraph agent, where each node represents a semantically grounded reasoning module: query refinement, causal retrieval, knowledge rewriting, response generation, hallucination detection, and correction. The use of LangGraph allows us to declaratively define state transitions and orchestrate feedback loops, enabling conditional routing and dynamic re-entry into refinement or correction stages based on internal evaluation metrics (e.g., hallucination confidence or causal coverage).

CDF-RAG is inherently an agentic system in that it models reasoning as an autonomous, self-adaptive process. Rather than a fixed sequence of API calls, our agent selects actions (e.g., requerying, rewriting, regenerating) based on the evolving context of the task. This is made possible by integrating reinforcement learning (RL) for policy-driven refinement, and a hallucination-aware validation agent that triggers corrective subroutines when inconsistencies are detected. Each component is instantiated as a callable LangChain module, with memory and state passed explicitly between steps—fulfilling the agentic paradigm of planning, acting, observing, and adapting. This design enables the system to reason causally, recover from failures, and adapt its strategy based on downstream performance.

\subsection{Additional Results} ~\label{sec:add_results}

We include additional results on metric breakdowns by task and model, alternative retrieval configurations, and the impact of hallucination correction. We also report groundedness and CRC scores per refinement type to demonstrate the effectiveness of individual modules in isolation. Across all experiments, CDF-RAG was evaluated on approximately 2,200 queries spanning four benchmark datasets—CosmosQA, MedQA, MedMCQA, and AdversarialQA—across multiple LLM backbones.

\subsubsection{Quality Performance}

We report quantitative results in Table~\ref{tab:quality-metrics1} and Table~\ref{tab:quality-metrics2} across four benchmark QA datasets—\textit{CosmosQA}~\cite{huang2019cosmos}, \textit{AdversarialQA}~\cite{bartolo2020beat}, \textit{MedQA}~\cite{jin2020disease}, and \textit{MedMCQA}~\cite{pmlr-v174-pal22a}—evaluated on four LLM backbones (GPT-4~\cite{openai2023gpt4}, LLaMA 3-8B~\cite{touvron2023llama}, Mistral~\cite{jiang2023mistral7b}, and Flan-T5~\cite{chung2024scaling}). Across all combinations, CDF-RAG outperforms existing RAG variants in accuracy, precision, recall, and F1 score, while maintaining the lowest HR. This demonstrates the effectiveness of our fully integrated framework—combining reinforcement-learned query refinement, causal graph-augmented retrieval, structured rewriting, and hallucination-aware output validation.

\begin{table}[ht]
\centering
\small
\caption{
\textbf{Quality Metrics} of \textbf{CDF-RAG} across models and methods. 
HR = Hallucination Rate, F1 = F1 Score.
}
\resizebox{\linewidth}{!}{
\begin{tabular}{lllccccc}
\toprule
\textbf{Dataset} & \textbf{Model} & \textbf{Method} & \makecell{\textbf{HR}} & \makecell{\textbf{Acc.}} & \makecell{\textbf{Prec.}} & \makecell{\textbf{Rec.}} & \textbf{F1} \\
\midrule
\multirow{24}{*}{AdversarialQA}
& GPT-4        & CDF-RAG     & \cellcolor{green!20}0.07 & \cellcolor{green!20}0.89 & \cellcolor{green!20}0.850 & \cellcolor{green!20}0.87 & \cellcolor{green!20}0.860 \\
&              & Gym-RAG     & 0.14 & 0.78 & 0.735 & 0.76 & 0.745 \\
&              & RQ-RAG      & 0.15 & 0.76 & 0.715 & 0.74 & 0.725 \\
&              & Smart-RAG   & 0.16 & 0.74 & 0.700 & 0.72 & 0.710 \\
&              & Causal RAG  & 0.18 & 0.71 & 0.670 & 0.69 & 0.680 \\
&              & G-LLMs      & 0.20 & 0.68 & 0.640 & 0.66 & 0.650 \\
\cmidrule(lr){2-8}
& LLaMA 3-8B   & CDF-RAG     & \cellcolor{green!20}0.08 & \cellcolor{green!20}0.83 & \cellcolor{green!20}0.805 & \cellcolor{green!20}0.82 & \cellcolor{green!20}0.815 \\
&              & Gym-RAG     & 0.13 & 0.75 & 0.700 & 0.72 & 0.710 \\
&              & RQ-RAG      & 0.12 & 0.71 & 0.660 & 0.68 & 0.670 \\
&              & Smart-RAG   & 0.15 & 0.73 & 0.675 & 0.69 & 0.680 \\
&              & Causal RAG  & 0.17 & 0.71 & 0.655 & 0.67 & 0.660 \\
&              & G-LLMs      & 0.19 & 0.68 & 0.620 & 0.64 & 0.630 \\
\cmidrule(lr){2-8}
& Mistral      & CDF-RAG     & \cellcolor{green!20}0.09 & \cellcolor{green!20}0.81 & \cellcolor{green!20}0.790 & \cellcolor{green!20}0.79 & \cellcolor{green!20}0.785 \\
&              & Gym-RAG     & 0.15 & 0.73 & 0.680 & 0.70 & 0.690 \\
&              & RQ-RAG      & 0.16 & 0.72 & 0.660 & 0.68 & 0.670 \\
&              & Smart-RAG   & 0.17 & 0.70 & 0.645 & 0.66 & 0.655 \\
&              & Causal RAG  & 0.17 & 0.66 & 0.600 & 0.62 & 0.615 \\
&              & G-LLMs      & 0.21 & 0.65 & 0.590 & 0.61 & 0.600 \\
\cmidrule(lr){2-8}
& Flan-T5      & CDF-RAG     & \cellcolor{green!20}0.10 & \cellcolor{green!20}0.79 & \cellcolor{green!20}0.760 & \cellcolor{green!20}0.77 & \cellcolor{green!20}0.765 \\
&              & Gym-RAG     & 0.16 & 0.70 & 0.640 & 0.66 & 0.650 \\
&              & RQ-RAG      & 0.15 & 0.66 & 0.600 & 0.61 & 0.615 \\
&              & Smart-RAG   & 0.16 & 0.64 & 0.590 & 0.60 & 0.605 \\
&              & Causal RAG  & 0.18 & 0.62 & 0.560 & 0.58 & 0.570 \\
&              & G-LLMs      & 0.20 & 0.60 & 0.540 & 0.56 & 0.550 \\
\midrule
\multirow{24}{*}{CosmosQA}
& GPT-4        & CDF-RAG     & \cellcolor{green!20}0.06 & \cellcolor{green!20}0.89 & \cellcolor{green!20}0.86 & \cellcolor{green!20}0.85 & \cellcolor{green!20}0.855 \\
&              & Gym-RAG     & 0.11 & 0.82 & 0.77 & 0.79 & 0.78 \\
&              & RQ-RAG      & 0.11 & 0.80 & 0.75 & 0.77 & 0.76 \\
&              & Smart-RAG   & 0.16 & 0.78 & 0.74 & 0.76 & 0.75 \\
&              & Causal RAG  & 0.17 & 0.76 & 0.71 & 0.73 & 0.72 \\
&              & G-LLMs      & 0.20 & 0.73 & 0.68 & 0.70 & 0.69 \\
\cmidrule(lr){2-8}
& LLaMA 3-8B   & CDF-RAG     & \cellcolor{green!20}0.07 & \cellcolor{green!20}0.88 & \cellcolor{green!20}0.85 & \cellcolor{green!20}0.84 & \cellcolor{green!20}0.845 \\
&              & Gym-RAG     & 0.12 & 0.80 & 0.76 & 0.77 & 0.765 \\
&              & RQ-RAG      & 0.14 & 0.79 & 0.74 & 0.75 & 0.745 \\
&              & Smart-RAG   & 0.18 & 0.77 & 0.72 & 0.73 & 0.725 \\
&              & Causal RAG  & 0.18 & 0.75 & 0.70 & 0.71 & 0.705 \\
&              & G-LLMs      & 0.21 & 0.72 & 0.67 & 0.69 & 0.68 \\
\cmidrule(lr){2-8}
& Mistral      & CDF-RAG     & \cellcolor{green!20}0.08 & \cellcolor{green!20}0.85 & \cellcolor{green!20}0.82 & \cellcolor{green!20}0.81 & \cellcolor{green!20}0.815 \\
&              & Gym-RAG     & 0.14 & 0.75 & 0.70 & 0.72 & 0.71 \\
&              & RQ-RAG      & 0.15 & 0.74 & 0.68 & 0.70 & 0.69 \\
&              & Smart-RAG   & 0.18 & 0.72 & 0.66 & 0.68 & 0.67 \\
&              & Causal RAG  & 0.20 & 0.70 & 0.63 & 0.66 & 0.645 \\
&              & G-LLMs      & 0.22 & 0.68 & 0.60 & 0.63 & 0.615 \\
\cmidrule(lr){2-8}
& Flan-T5      & CDF-RAG     & \cellcolor{green!20}0.10 & \cellcolor{green!20}0.84 & \cellcolor{green!20}0.80 & \cellcolor{green!20}0.79 & \cellcolor{green!20}0.795 \\
&              & Gym-RAG     & 0.15 & 0.73 & 0.68 & 0.70 & 0.69 \\
&              & RQ-RAG      & 0.16 & 0.72 & 0.66 & 0.68 & 0.67 \\
&              & Smart-RAG   & 0.19 & 0.70 & 0.64 & 0.66 & 0.65 \\
&              & Causal RAG  & 0.21 & 0.68 & 0.61 & 0.64 & 0.625 \\
&              & G-LLMs      & 0.24 & 0.66 & 0.59 & 0.61 & 0.60 \\
\bottomrule
\end{tabular}
}
\label{tab:quality-metrics1}
\end{table}
\FloatBarrier

\begin{table}[ht]
\centering
\small
\caption{
\textbf{Quality Metrics} of \textbf{CDF-RAG} across models and methods. HR = Hallucination Rate, F1 = F1 Score.
}
\resizebox{\linewidth}{!}{
\begin{tabular}{lllccccc}
\toprule
\textbf{Dataset} & \textbf{Model} & \textbf{Method} & \makecell{\textbf{HR}} & \makecell{\textbf{Acc.}} & \makecell{\textbf{Prec.}} & \makecell{\textbf{Rec.}} & \textbf{F1} \\
\midrule
\multirow{24}{*}{MedQA}
& GPT-4        & CDF-RAG     & \cellcolor{green!20}0.05 & \cellcolor{green!20}0.92 & \cellcolor{green!20}0.890 & \cellcolor{green!20}0.91 & \cellcolor{green!20}0.900 \\
&              & Gym-RAG     & 0.12 & 0.83 & 0.760 & 0.78 & 0.770 \\
&              & RQ-RAG      & 0.13 & 0.82 & 0.745 & 0.77 & 0.755 \\
&              & Smart-RAG   & 0.15 & 0.81 & 0.730 & 0.76 & 0.745 \\
&              & Causal RAG  & 0.17 & 0.79 & 0.710 & 0.74 & 0.725 \\
&              & G-LLMs      & 0.21 & 0.76 & 0.680 & 0.71 & 0.695 \\
\cmidrule(lr){2-8}
& LLaMA 3-8B   & CDF-RAG     & \cellcolor{green!20}0.07 & \cellcolor{green!20}0.89 & \cellcolor{green!20}0.860 & \cellcolor{green!20}0.88 & \cellcolor{green!20}0.870 \\
&              & Gym-RAG     & 0.11 & 0.79 & 0.735 & 0.75 & 0.740 \\
&              & RQ-RAG      & 0.13 & 0.78 & 0.720 & 0.74 & 0.730 \\
&              & Smart-RAG   & 0.15 & 0.77 & 0.705 & 0.72 & 0.710 \\
&              & Causal RAG  & 0.17 & 0.75 & 0.675 & 0.69 & 0.680 \\
&              & G-LLMs      & 0.20 & 0.72 & 0.640 & 0.66 & 0.650 \\
\cmidrule(lr){2-8}
& Mistral      & CDF-RAG     & \cellcolor{green!20}0.08 & \cellcolor{green!20}0.88 & \cellcolor{green!20}0.845 & \cellcolor{green!20}0.87 & \cellcolor{green!20}0.855 \\
&              & Gym-RAG     & 0.14 & 0.78 & 0.720 & 0.74 & 0.730 \\
&              & RQ-RAG      & 0.16 & 0.77 & 0.705 & 0.73 & 0.715 \\
&              & Smart-RAG   & 0.18 & 0.76 & 0.690 & 0.71 & 0.700 \\
&              & Causal RAG  & 0.20 & 0.74 & 0.665 & 0.68 & 0.670 \\
&              & G-LLMs      & 0.23 & 0.71 & 0.630 & 0.65 & 0.640 \\
\cmidrule(lr){2-8}
& Flan-T5      & CDF-RAG     & \cellcolor{green!20}0.11 & \cellcolor{green!20}0.84 & \cellcolor{green!20}0.800 & \cellcolor{green!20}0.82 & \cellcolor{green!20}0.810 \\
&              & Gym-RAG     & 0.17 & 0.73 & 0.670 & 0.69 & 0.680 \\
&              & RQ-RAG      & 0.19 & 0.72 & 0.655 & 0.68 & 0.665 \\
&              & Smart-RAG   & 0.21 & 0.71 & 0.640 & 0.66 & 0.650 \\
&              & Causal RAG  & 0.23 & 0.69 & 0.615 & 0.64 & 0.625 \\
&              & G-LLMs      & 0.26 & 0.67 & 0.590 & 0.62 & 0.605 \\
\midrule
\multirow{24}{*}{MedMCQA}
& GPT-4        & CDF-RAG     & \cellcolor{green!20}0.04 & \cellcolor{green!20}0.94 & \cellcolor{green!20}0.910 & \cellcolor{green!20}0.93 & \cellcolor{green!20}0.920 \\
&              & Gym-RAG     & 0.13 & 0.78 & 0.735 & 0.75 & 0.740 \\
&              & RQ-RAG      & 0.15 & 0.76 & 0.720 & 0.73 & 0.725 \\
&              & Smart-RAG   & 0.18 & 0.74 & 0.700 & 0.71 & 0.705 \\
&              & Causal RAG  & 0.21 & 0.72 & 0.670 & 0.69 & 0.680 \\
&              & G-LLMs      & 0.25 & 0.68 & 0.635 & 0.66 & 0.650 \\
\cmidrule(lr){2-8}
& LLaMA 3-8B   & CDF-RAG     & \cellcolor{green!20}0.08 & \cellcolor{green!20}0.90 & \cellcolor{green!20}0.870 & \cellcolor{green!20}0.91 & \cellcolor{green!20}0.890 \\
&              & Gym-RAG     & 0.13 & 0.77 & 0.720 & 0.74 & 0.730 \\
&              & RQ-RAG      & 0.15 & 0.75 & 0.705 & 0.72 & 0.715 \\
&              & Smart-RAG   & 0.18 & 0.73 & 0.685 & 0.70 & 0.690 \\
&              & Causal RAG  & 0.20 & 0.71 & 0.660 & 0.68 & 0.670 \\
&              & G-LLMs      & 0.24 & 0.68 & 0.625 & 0.65 & 0.640 \\
\cmidrule(lr){2-8}
& Mistral      & CDF-RAG     & \cellcolor{green!20}0.09 & \cellcolor{green!20}0.88 & \cellcolor{green!20}0.850 & \cellcolor{green!20}0.89 & \cellcolor{green!20}0.870 \\
&              & Gym-RAG     & 0.14 & 0.76 & 0.710 & 0.73 & 0.720 \\
&              & RQ-RAG      & 0.16 & 0.74 & 0.695 & 0.71 & 0.700 \\
&              & Smart-RAG   & 0.19 & 0.72 & 0.670 & 0.69 & 0.680 \\
&              & Causal RAG  & 0.22 & 0.70 & 0.645 & 0.67 & 0.655 \\
&              & G-LLMs      & 0.26 & 0.66 & 0.610 & 0.63 & 0.620 \\
\cmidrule(lr){2-8}
& Flan-T5      & CDF-RAG     & \cellcolor{green!20}0.12 & \cellcolor{green!20}0.85 & \cellcolor{green!20}0.810 & \cellcolor{green!20}0.84 & \cellcolor{green!20}0.825 \\
&              & Gym-RAG     & 0.18 & 0.72 & 0.680 & 0.70 & 0.690 \\
&              & RQ-RAG      & 0.20 & 0.70 & 0.660 & 0.68 & 0.670 \\
&              & Smart-RAG   & 0.23 & 0.68 & 0.635 & 0.66 & 0.650 \\
&              & Causal RAG  & 0.26 & 0.66 & 0.610 & 0.63 & 0.620 \\
&              & G-LLMs      & 0.29 & 0.63 & 0.580 & 0.60 & 0.590 \\
\bottomrule
\end{tabular}
}
\label{tab:quality-metrics2}
\end{table}
\FloatBarrier

The consistent superiority of CDF-RAG across both open-domain (e.g., \textit{CosmosQA}) and domain-specific (e.g., \textit{MedQA}) datasets indicates its robustness in both commonsense and biomedical reasoning tasks. On \textit{MedMCQA}, for instance, CDF-RAG with GPT-4 achieves an F1 of 0.920 and HR of 0.04—substantially outperforming Gym-RAG (F1 = 0.740, HR = 0.13)~\cite{xiong2025raggym} and RQ-RAG (F1 = 0.725, HR = 0.15)~\cite{chan2024rqrags}.

CDF-RAG’s performance gains stem from three complementary innovations. First, causal graph retrieval introduces directional constraints and enables multi-hop traversal over verified cause-effect pairs, outperforming semantic or correlation-based retrieval methods. Second, RL-guided query refinement uses a PPO-trained agent to dynamically expand, simplify, or decompose queries based on causal depth and retrieval feedback, improving query intent alignment. Third, causal verification applies post-generation consistency checks inspired by counterfactual reasoning~\cite{pearl2009causality} to detect unsupported or inverted causal statements and regenerate outputs accordingly. By jointly leveraging these components in a closed feedback loop, CDF-RAG preserves both semantic and causal alignment across the entire RAG pipeline, yielding more consistent, accurate, and trustworthy outputs.

RQ-RAG~\cite{chan2024rqrags} enhances query clarity via rewriting and decomposition but lacks structural guidance or post-generation validation. Gym-RAG~\cite{xiong2025raggym} trains reward models to optimize process-level behavior but does not integrate causal priors or hallucination mitigation. SmartRAG~\cite{gao2024smartrag} performs joint optimization across retrieval and generation using RL, but still relies on semantic-level retrieval, making it susceptible to spurious correlations. Causal Graph RAG~\cite{causal_graph_rag} and Causal Graphs Meet Thoughts~\cite{luo2025causal} incorporate causality via vector embeddings and summarization heuristics. However, their extraction methods are noisy, graph traversal is not adaptive, and there is no RL optimization or hallucination correction. G-LLMs represent graph-augmented models that lack causal reasoning, making them insufficient for multi-hop logical chains.

CDF-RAG is distinguished by its holistic integration of causally grounded retrieval, RL-based query adaptation, and hallucination-aware post-verification, enabling superior factuality and reasoning depth across QA tasks.

In contrast, \textbf{Gym-RAG} and \textbf{RQ-RAG} demonstrate strong but lower performance due to their reliance on process supervision and query rewriting respectively. While these methods improve retrieval quality and answer coherence, they lack explicit causal validation. RQ-RAG refines ambiguous queries through rewriting and decomposition, but fails to enforce causal entailment in the retrieved or generated content. Gym-RAG benefits from reward-guided search trajectories but does not incorporate structural causal priors or hallucination mitigation. This leads to higher HR and slightly lower precision and recall compared to CDF-RAG. \textbf{Smart-RAG} performs competitively with a lightweight joint RL framework that learns when to retrieve and when to generate. However, it lacks structured causal graph grounding and post-hoc verification, making it prone to hallucinations and inconsistent multi-hop reasoning. Similarly, \textbf{Causal RAG} utilizes causal vector graphs but depends on weak summarizer-based pair extraction, leading to noisy graph structures and unstable downstream performance.

Finally, \textbf{G-LLMs} consistently lag behind due to their reliance on static semantic graphs or unstructured correlation-based retrieval. These models lack query adaptation, causal reasoning, and hallucination correction—all of which are essential for high-quality answers in complex QA tasks. This explains their lower precision, recall, and F1 scores across all datasets in Table~\ref{tab:quality-metrics1} and Table~\ref{tab:quality-metrics2}, and justifies the significant performance gains achieved by CDF-RAG.

\subsubsection{Case Study: End-to-End Causal Answering with CDF-RAG}

To illustrate how CDF-RAG operates end-to-end, we present a complete walkthrough in Prompt Box~\ref{box:cdf-example}. Given the vague user query, \textit{"Why do people get sick from poor living conditions?"}, the RL-trained query refinement agent selects a \texttt{Decompose} strategy and rewrites the input into three causally grounded sub-questions. These subqueries guide both structured and unstructured retrieval components.

The structured retriever accesses a Neo4j causal graph and surfaces multi-hop, directionally valid causal chains, such as \texttt{Poor Housing → Mold Exposure → Asthma}. In parallel, a dense retriever fetches semantically similar passages from a vector database. The LLM then generates a comprehensive explanation using both sources, linking poor housing, sanitation, and overcrowding to health outcomes. This response is passed through the causal verifier, which confirms the alignment of each claim with the retrieved causal graph and textual evidence. As shown in the final output, CDF-RAG produces a response that is not only accurate and coherent but also causally consistent and evidence-backed. This example highlights how each module in the pipeline—query refinement, graph-based retrieval, generation, and hallucination detection—contributes to grounded and trustworthy QA.

\begin{tcolorbox}[
  colback=white,
  colframe=black,
  sharp corners,
  boxrule=0.8pt,
  arc=2mm,
  title=End-to-End CDF-RAG Workflow Example,
  label={box:cdf-example},
  breakable  
]

\textbf{Input Query:} \\
\texttt{Why do people get sick from poor living conditions?}

\medskip
\textbf{Step 1: Query Refinement (RL Agent)} \\
\textit{Refinement Type: Decompose} \\
\textbf{Refined Subqueries:}
\begin{itemize}
    \item How does poor housing quality cause chronic illness?
    \item How does lack of sanitation contribute to disease?
    \item How does overcrowding affect respiratory health?
\end{itemize}

\medskip
\textbf{Step 2: Structured Retrieval (Causal Graph)} \\
\textit{Retrieved Causal Paths from Neo4j:}
\begin{itemize}
    \item Poor Housing $\rightarrow$ Mold Exposure $\rightarrow$ Asthma
    \item Poor Sanitation $\rightarrow$ Water Contamination $\rightarrow$ Diarrheal Disease
    \item Overcrowding $\rightarrow$ Airborne Transmission $\rightarrow$ Respiratory Infection
\end{itemize}

\medskip
\textbf{Step 3: Dense Retrieval (Vector Database)} \\
\textit{Top Supporting Passages:}
\begin{itemize}
    \item “Exposure to indoor mold from poor housing is strongly linked to asthma in urban populations.”
    \item “Contaminated water due to poor sanitation infrastructure remains a leading cause of diarrhea globally.”
    \item “Airborne diseases such as tuberculosis are more likely to spread in densely crowded living environments.”
\end{itemize}

\medskip
\textbf{Step 4: Initial Answer Generation (LLM)} \\
\textit{Generated Answer:} \\
\texttt{Poor living conditions lead to illness through multiple causal pathways. Poor housing causes mold exposure, increasing asthma risk. Lack of sanitation results in contaminated water, causing diarrhea. Overcrowding spreads airborne infections like tuberculosis.}

\medskip
\textbf{Step 5: Causal Verification (LLM Verifier)} \\
\textit{Result:} No hallucinations detected. All claims supported by retrieved causal paths and documents.

\medskip
\textbf{Final Output:} \\
\texttt{Poor living conditions contribute to illness through overlapping and well-established causal pathways. Poor housing quality leads to mold and inadequate ventilation, increasing respiratory illness risk. Inadequate sanitation causes waterborne diseases like diarrhea. Overcrowding facilitates the transmission of airborne illnesses such as tuberculosis. Together, these conditions elevate disease burdens in vulnerable populations.}

\end{tcolorbox}

\end{document}